\documentclass[10pt,twocolumn,letterpaper]{article}

\usepackage[pagenumbers]{cvpr} 
\usepackage{bm}
\usepackage{multirow}
\usepackage{makecell}

\usepackage[accsupp]{axessibility}  

\definecolor{cvprblue}{rgb}{0.21,0.49,0.74}
\usepackage[pagebackref,breaklinks,colorlinks,allcolors=cvprblue]{hyperref}

\def\paperID{4294} 
\def\confName{CVPR}
\def\confYear{2025}

\title{Diffusion-4K: Ultra-High-Resolution Image Synthesis with Latent Diffusion Models}

\author{Jinjin Zhang\textsuperscript{1,2} \hspace{0.5em} 
Qiuyu Huang\textsuperscript{3} \hspace{0.5em}
Junjie Liu\textsuperscript{3} \hspace{0.5em} 
Xiefan Guo\textsuperscript{1,2} \hspace{0.5em} 
Di Huang\textsuperscript{1,2}\footnotemark[1]  \\
\textsuperscript{1}State Key Laboratory of Complex and Critical Software Environment, \\Beihang University, Beijing 100191, China\\
\textsuperscript{2}School of Computer Science and Engineering, Beihang University, Beijing 100191, China\\
\textsuperscript{3}Meituan\\
{\tt\small{\{jinjin.zhang, xfguo, dhuang\}@buaa.edu.cn}     \{huangqiuyu, liujunjie10\}@meituan.com} 
}

\begin{document}

\twocolumn[{
\renewcommand\twocolumn[1][]{#1}
\maketitle
\vspace{-20pt}
\begin{center}
\centering
\captionsetup{type=figure}
\includegraphics[width=0.95\linewidth]{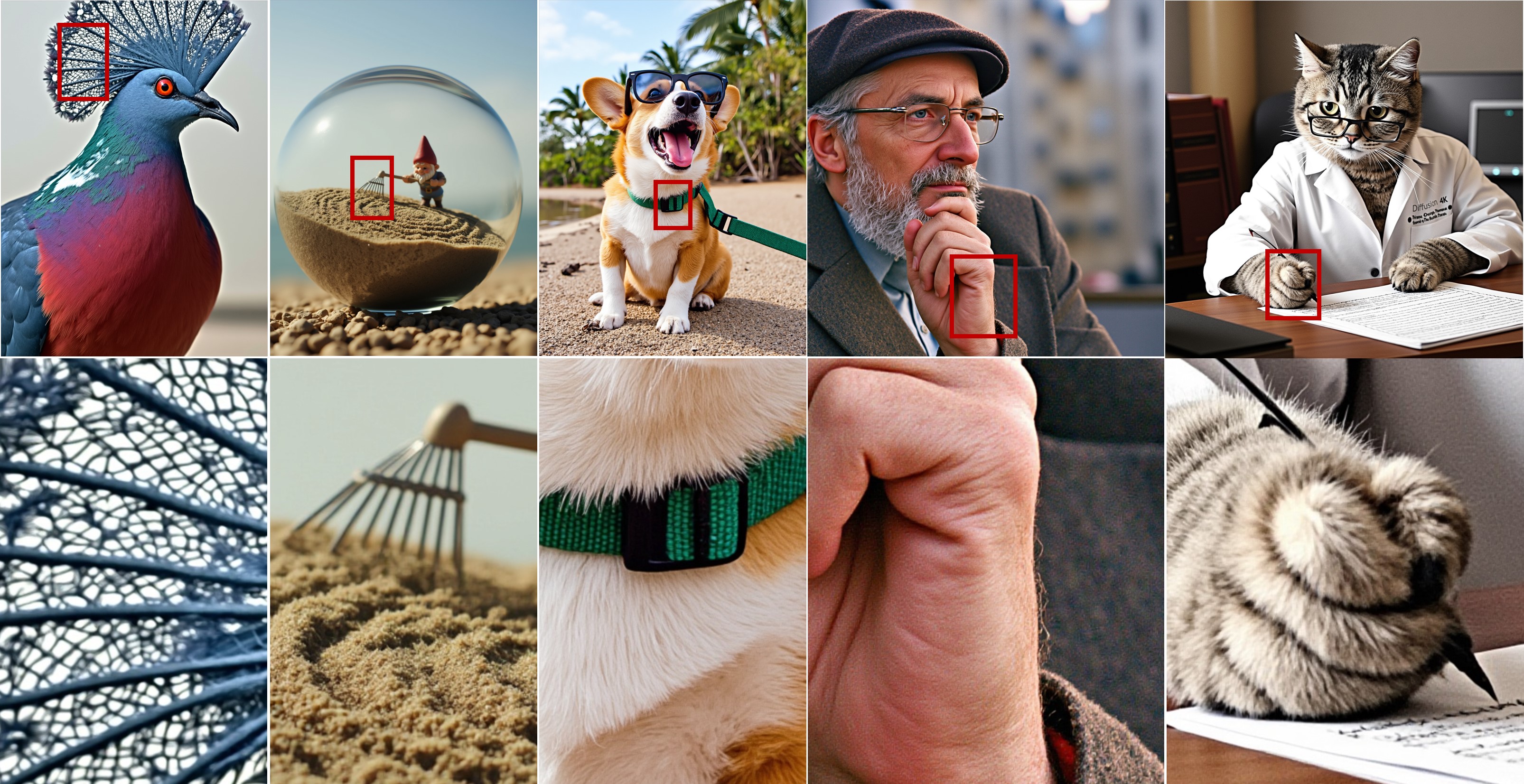}
\captionof{figure}{Example results synthesized by our \textbf{Diffusion-4K}, emphasizing exceptional fine details in generated 4K images.}
\label{fig:demo}
\end{center}
}]

\footnotetext[1]{Corresponding author.}

\begin{abstract}

In this paper, we present Diffusion-4K, a novel framework for direct ultra-high-resolution image synthesis using text-to-image diffusion models.
The core advancements include:
(1) Aesthetic-4K Benchmark: addressing the absence of a publicly available 4K image synthesis dataset, we construct Aesthetic-4K, a comprehensive benchmark for ultra-high-resolution image generation. 
We curated a high-quality 4K dataset with carefully selected images and captions generated by GPT-4o.
Additionally, we introduce GLCM Score and Compression Ratio metrics to evaluate fine details, combined with holistic measures such as FID, Aesthetics and CLIPScore for a comprehensive assessment of ultra-high-resolution images.
(2) Wavelet-based Fine-tuning: we propose a wavelet-based fine-tuning approach for direct training with photorealistic 4K images, applicable to various latent diffusion models, demonstrating its effectiveness in synthesizing highly detailed 4K images.
Consequently, Diffusion-4K achieves impressive performance in high-quality image synthesis and text prompt adherence, especially when powered by modern large-scale diffusion models (e.g., SD3-2B and Flux-12B).
Extensive experimental results from our benchmark demonstrate the superiority of Diffusion-4K in ultra-high-resolution image synthesis.
Code is available at \href{https://github.com/zhang0jhon/diffusion-4k}{https://github.com/zhang0jhon/diffusion-4k}.

\end{abstract}

\section{Introduction}

Diffusion models have proven to be highly effective for modeling high-dimensional, perceptual data such as images~\cite{ho2020denoising, song2020score, nichol2021improved, karras2022elucidating, peebles2023scalable}.
In recent years, latent diffusion models have made significant advancements in text-to-image synthesis, demonstrating impressive generalization capabilities at high resolutions~\cite{rombach2022high, podell2023sdxl, esser2024scaling, chen2024pixart, li2024playground, liu2024playground}. 
Replacing convolutional U-Net with a transformer architecture shows promising improvements as model scalability increases, \eg, Stable Diffusion 3 (SD3) at 8B~\cite{esser2024scaling}, Flux at 12B~\cite{Flux:2024:Online}, and Playground v3 at 24B~\cite{liu2024playground}.
Alternatively, flow-based formulations, using data or velocity prediction, have emerged as an viable choice due to their faster convergence and enhanced performance~\cite{liu2022flow, karras2022elucidating, albergo2022building, lipman2022flow, ma2024sit}.

Despite substantial progress, most latent diffusion models focus on training and generating images at $1024 \times 1024$, leaving direct ultra-high-resolution image synthesis largely unexplored.
Direct training and generating 4K images is valuable in realistic applications, but require considerable computational resources, especially as model parameters increase.
Concurrent approaches such as PixArt-$\Sigma$~\cite{chen2024pixart} and Sana~\cite{xie2024sana} address direct ultra-high-resolution image synthesis at 4K, showcasing the potentials of scalable latent diffusion transformer architectures.
Both PixArt-$\Sigma$ at 0.6B and Sana at 0.6/1.6B primarily focus on the efficiency of ultra-high-resolution image generation, however, the intrinsic advantages of 4K images, such as high-frequency details and rich textures, are neglected. 
Furthermore, due to the scarcity of photorealistic 4K images, there is currently no publicly available benchmark for 4K image synthesis, hindering further research on this valuable topic.

In this paper, we propose Diffusion-4K, a novel framework designed for direct ultra-high-resolution image synthesis using latent diffusion models.
Specifically, we introduce Aesthetic-4K, including a high-quality dataset of curated ultra-high-resolution images, with corresponding captions generated by GPT-4o~\cite{hurst2024gpt}.
Additionally, most automated evaluation metrics, such as Fréchet Inception Distance (FID)~\cite{heusel2017gans}, Aesthetics~\cite{schuhmann2022laion} and CLIPScore~\cite{hessel2021clipscore},  offer only holistic measures at low resolutions, which are insufficient for the comprehensive benchmarking in high-resolution image synthesis, particularly for 4K images.
To address the limitations, we introduce Gray Level Co-occurrence Matrix (GLCM) Score and Compression Ratio metrics, focusing on assessment of fine details in ultra-high-resolution images which has not yet been explored, with the goal of establishing a comprehensive benchmark for 4K image synthesis.
Additionally, we propose a wavelet-based fine-tuning method that emphasizes high-frequency components while preserving low-frequent approximation in ultra-high-resolution image synthesis.
Notably, our method is compatible with various latent diffusion models.
We conduct experiments with open-sourced latent diffusion models including SD3-2B~\cite{esser2024scaling} and Flux-12B~\cite{Flux:2024:Online}, capable of training and generating photorealistic images at 4096 $\times$ 4096 resolution.
Consequently, our method demonstrates superior performance in 4K image synthesis, underscoring the effectiveness of the Diffusion-4K framework.

The main contributions are summarized as follows:
\begin{itemize}
    \item We establish Aesthetic-4k, a comprehensive benchmark for 4K image synthesis, including a high-quality dataset of ultra-high-resolution images with corresponding precise captions, and elaborated evaluation metrics for ultra-high-resolution images generation.
    \item We propose a wavelet-based fine-tuning approach for latent diffusion model, focusing on generating ultra-high-resolution images with fine details.
    \item Extensive experiment results demonstrate the effectiveness and generalization of our proposed method in 4K image synthesis, particularly when powered by large-scale diffusion transformers, \eg SD3 and Flux.
\end{itemize}

\section{Related work}

\noindent\textbf{Latent Diffusion Models.} 
Stable Diffusion (SD)~\cite{rombach2022high} proposes latent diffusion models to operate diffusion process in compressed latent space using Variational Auto-Encoder (VAE).
Widely used VAEs~\cite{peebles2023scalable, esser2024scaling, chen2023pixart} employ a down-sampling factor of $F=8$, compressing pixel space $\mathbb{R}^{H \times W \times 3}$ into latent space $\mathbb{R}^{\frac{H}{F} \times \frac{W}{F} \times C}$, where $H$ and $W$ represent height and width respectively, and $C$ denotes the channel of latent space.
In the field of latent diffusion models, Diffusion Transformer (DiT)~\cite{peebles2023scalable} has made significant progress in the past year.
Typically, the patch size of DiT is set to $P=2$, resulting in $\frac{H}{FP} \times \frac{H}{FP}$ tokens.
The transformer architecture demonstrates excellent scalability in latent diffusion models, as evidenced by PixArt~\cite{chen2023pixart, chen2024pixart}, SD3~\cite{esser2024scaling}, Flux~\cite{Flux:2024:Online}, Playground~\cite{li2024playground, liu2024playground}, \etc.

In text-to-image synthesis, the text encoder plays a crucial role in prompt coherence.
SD employs CLIP~\cite{radford2021learning} as its text encoder, while subsequent diffusion models, such as Imagen~\cite{saharia2022photorealistic} and PixArt~\cite{chen2023pixart, chen2024pixart}, utilize T5-XXL~\cite{raffel2020exploring} for text feature extraction.
Recent approaches, such as SD3~\cite{esser2024scaling} and Flux~\cite{Flux:2024:Online}, combine both CLIP and T5-XXL for enhanced text understanding. 
DALL-E 3~\cite{betker2023improving} demonstrates that training with descriptive image captions can significantly enhance the prompt coherence of text-to-image models.
Sana~\cite{xie2024sana} utilizes the latest decoder-only Large Language
Model (LLM), Gemma 2~\cite{team2024gemma}, as its text encoder to improve understanding and reasoning capabilities related to text prompts.

\noindent\textbf{High-Resolution Image Synthesis.}  
High-resolution image generation is valuable in various practical applications such as industry and entertainment~\cite{esser2021taming, chai2022any}.  
Currently, advanced latent diffusion models typically train and synthesize images at $1024 \times 1024$ for high-resolution image generation due to computational complexity ~\cite{rombach2022high, ramesh2021zero, ramesh2022hierarchical, podell2023sdxl, sauer2023adversarial, sauer2024fast}. 
However, increasing image resolution introduces quadratic computational costs, posing challenges particularly for 4K image synthesis. 
Several training-free fusion approaches for 4K image generation have been developed based on existing latent diffusion models~\cite{bar2023multidiffusion, du2024demofusion}. 
Additionally, Stable Cascade~\cite{pernias2023wurstchen} employs multiple diffusion networks to increase resolution progressively. 
However, such ensemble methods can introduce cumulative errors. 

PixArt-$\Sigma$~\cite{chen2024pixart} pioneers to direct image generation close to 4K ($3840 \times 2160$) with efficient token compression for DiT, significantly improving efficiency and enabling ultra-high-resolution image generation.
Sana~\cite{xie2024sana}, a pipeline designed to efficiently and cost-effectively train and synthesize 4K images, is capable of generating images at resolutions ranging from $1024 \times 1024$ to $4096 \times 4096$.
Sana introduces a deep compression VAE~\cite{chen2024deep} that compresses images with an aggressive down-sampling factor of $F=32$, enabling content creation at low cost.
Despite significant improvements in resolution, both PixArt-$\Sigma$-0.6B and Sana-0.6B/1.6B ignore the intrinsic high-frequency details and rich textures in 4K image synthesis.
Furthermore, the potential for scalable DiT in 4K image generation remains unexplored.

\section{Methods}

In this section, we elaborate on the details of Diffusion-4K, illustrating how it facilitates photorealistic 4K image synthesis. 
In \cref{sec:aesthetic-4k}, we introduce Aesthetic-4K, outlining the design principles and the process of constructing our human-centric benchmark with ultra-high-resolution images.
Subsequently, we present a wavelet-based fine-tuning approach for direct training with photorealistic 4K images, applicable to various latent diffusion models, in \cref{sec:wavelet}.

\subsection{Aesthetic-4K Benchmark}
\label{sec:aesthetic-4k}

Existing benchmarks for advanced latent diffusion models generally conduct experiments with images at $1024 \times 1024$ resolution~\cite{podell2023sdxl, li2024playground, chen2023pixart, esser2024scaling}, however, direct training and assessing with 4K images has not been thoroughly addressed yet.
In this section, we construct Aesthetic-4K, a comprehensive benchmark for ultra-high-resolution image synthesis.
The details are outlined as follows.

\noindent\textbf{Design Principles.}
\emph{Emphasis on human-centric perceptual cognition:} To design a benchmark centered on human-level perceptual cognition, our primary goal is to explore the key factors of human perception on 4K images and formulate quantifiable indicators, while evaluating the general abilities of the generative model in these key factors. 
However, most automated evaluation metrics, such as FID~\cite{heusel2017gans}, Aesthetics~\cite{schuhmann2022laion}, and CLIPScore~\cite{hessel2021clipscore}, provide only holistic measures at lower resolutions, which are insufficient for comprehensive assessment, particularly lacking the ability of evaluating fine details in 4K images.

\begin{figure}
  \centering
  \includegraphics[width=\linewidth]{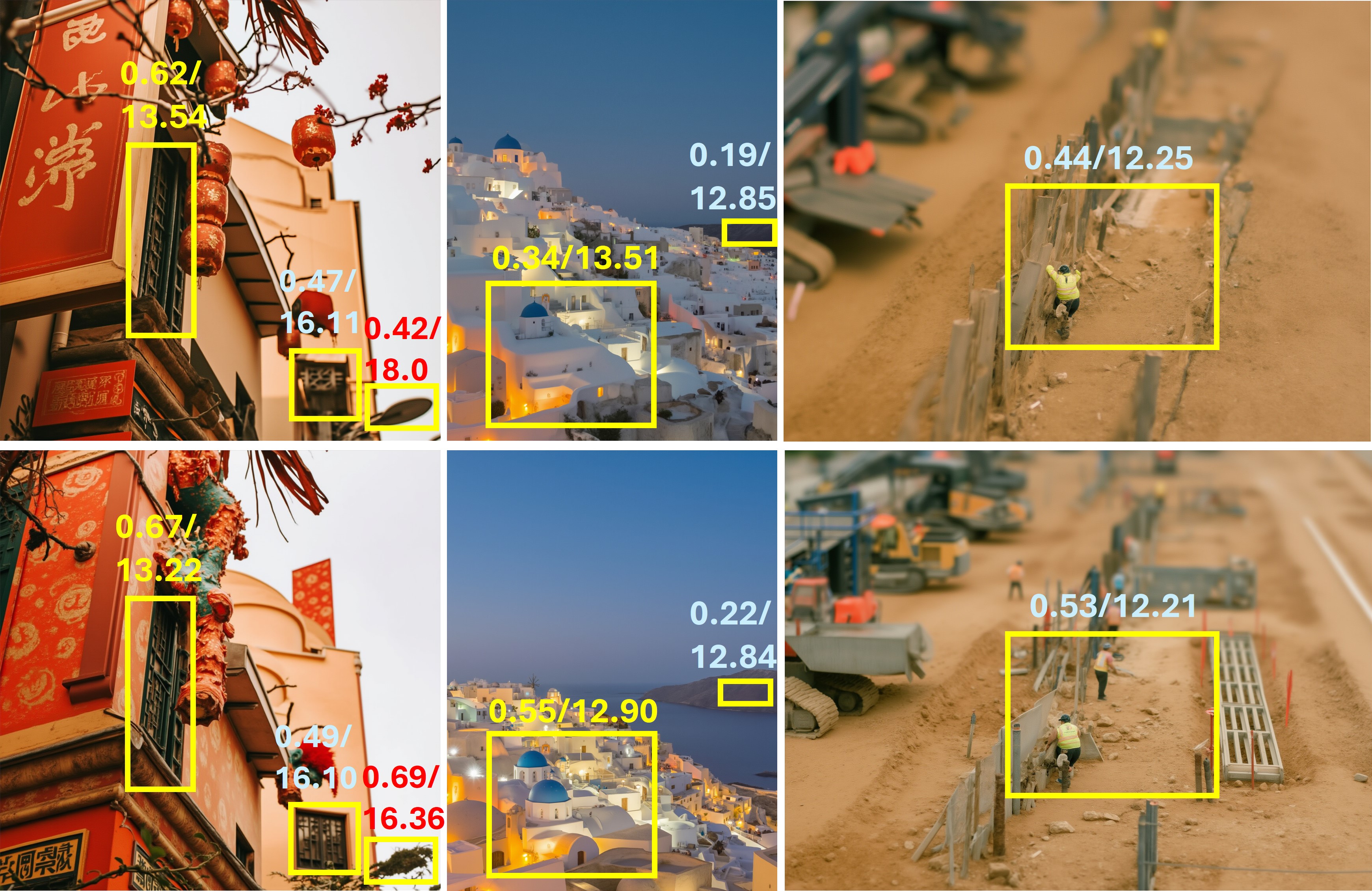}
  \caption{Analysis of GLCM Score$\uparrow$ / Compression Ratio$\downarrow$. Our indicators demonstrate strong alignment with human-centric perceptual cognition at the level of local patches. }
  \label{fig:image_align_with_glcm}
\end{figure}

Inspired by relevant literature in perceptual psychology, we found that highly structured textures play a decisive role in human perceptual cognition~\cite{julesz1981textons, stockwell2020texture, bergen1988early}. 
Therefore, we proposed additional indicators that accurately measure the richness of highly structured textures, including the commonly used quantitative indicators for texture analysis GLCM and the image compression ratio with discrete cosine transform (DCT). 
Considering human visual sensitivities, the GLCM, which captures the varying textural patterns, optical flow, and distortion through comprehensive spatial interactions of neighboring pixels, enables the creation of a truly representative characterization of human visual perception~\cite{gadkari2004image}.
Simultaneously, the image compression ratio with DCT serves as an important reference for assessing the preservation of fine details in 4K images.
To demonstrate alignment with human preferences, we conduct quantitative analysis of our indicators with adequate images, and present examples in \cref{fig:image_align_with_glcm}.
Moreover, as depicted in \cref{tab:metrics}, we compute the Spearman Rank-order Correlation Coefficient (SRCC) and Pearson Linear Correlation Coefficient (PLCC) with  human evaluations using various patches sampled from generated images, demonstrating the superior alignment with human ratings compared to no-reference image quality assessment metrics such as MUSIQ~\cite{ke2021musiq} and MANIQA~\cite{yang2022maniqa}. 
Our benchmark design ensures that the model performance assessment is directly relevant to human decision-making and cognitive abilities.

\begin{table}[t]
\centering
\resizebox{0.98\columnwidth}{!}{
\begin{tabular}{c|c|c|c|c}
\toprule
Metric & GLCM Score & Compression Ratio & MUSIQ & MANIQA \\
\midrule
SRCC  & 0.75 & 0.53 & 0.36 & 0.20 \\
\midrule
PLCC  & 0.77 & 0.56 & 0.41 & 0.26 \\
\bottomrule
\end{tabular}
}
\caption{Correlation with human evaluation. } 
\label{tab:metrics}
\end{table}

\begin{figure*}[t]
  \centering
  \begin{subfigure}{0.49\linewidth}
    \includegraphics[width=\linewidth]{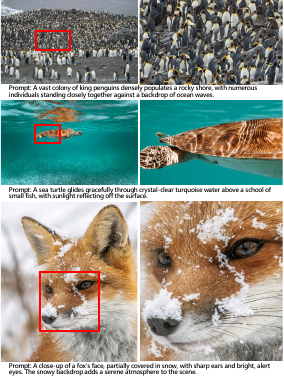}
    \caption{Image-text samples in training set.}
    \label{fig:short-a}
  \end{subfigure}
  \hfill
  \begin{subfigure}{0.49\linewidth}
    \includegraphics[width=\linewidth]{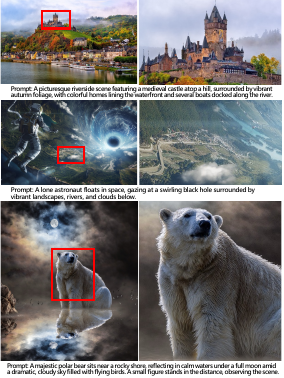}
    \caption{Image-text samples in evaluation set.}
    \label{fig:short-b}
  \end{subfigure}
  \caption{Illustration of image-text samples in the Aesthetic-4K dataset, which includes high-quality images and precise text prompts generated by GPT-4o, distinguished by high aesthetics and fine details. }
  \label{fig:aesthetic-4k}
\end{figure*}

\noindent\textbf{Aesthetic-4K Dataset.} Addressing the absence of a publicly available 4K dataset, we propose the Aesthetic-4K dataset that includes a well-curated training and evaluation set for in-depth study.
For the Aesthetic-4K training set, we collect high-quality images from the Internet, distinguished by exceptional aesthetics and fine details.
Notably, the Aesthetic-4K training set consists of 12,015 high-quality images with a median height of $4128$ and median width of $4640$, representing a significant improvement in training data for ultra-high-resolution image synthesis.
Additionally, we provide precise and detailed image captions generated by powerful GPT-4o~\cite{hurst2024gpt}, demonstrating strong alignment between vision and language. 
These curated images and corresponding captions comprise Aesthetic-Train, the training set of Aesthetic-4K.

For the evaluation set, Aesthetic-Eval, we select image-text pairs from the LAION-Aesthetics V2 6.5+ dataset, following the criterion of a short image side greater than 2048.
The LAION-Aesthetics dataset includes 625,000 image-text pairs with predicted aesthetic scores of 6.5 or higher in LAION-5B~\cite{schuhmann2022laion}.
Notably, the Aesthetic-4K evaluation set consists of 2,781 high-quality images with a median height of $2983$ and median width of $3613$. 
We do not sample the evaluation set from the collected images sourced from the Internet, to mitigate the risk of overfitting in comprehensive assessments.
Within the evaluation set, 195 images have a short side greater than 4096, which we refer to as Aesthetic-Eval@4096.
Instead of evaluating with images at $1024 \times 1024$ resolution~\cite{xie2024sana}, the proposed Aesthetic-4K evaluation set provides a suitable benchmark for ultra-high-resolution image synthesis.

In summary, the proposed dataset encompasses common categories in realistic scenarios, including nature, travel, fashion, animals, film, art, food, sports, street photography, \etc.
As illustrated in \cref{fig:aesthetic-4k}, we present several image-text pairs from both the training and evaluation sets of Aesthetic-4K, clearly demonstrating their exceptional quality.
For more details about the Aesthetic-4K dataset, please refer to the supplementary materials.


\noindent\textbf{Comprehensive Evaluation Metrics.} 
To construct a comprehensive human-centric benchmark for 4K image synthesis, we provide conventional evaluation metrics in generative models, such as FID~\cite{heusel2017gans}, Aesthetics~\cite{schuhmann2022laion}, and CLIPScore~\cite{hessel2021clipscore}, for an intuitive understanding of image synthesis from a holistic perspective. 

In addition, we report quantitative results for the proposed metrics assessing fine details in 4K images, including the GLCM Score and Compression Ratio, as significant complements to human-centric perceptual cognition in 4K images, which have not been addressed previously.
The GLCM Score is formulated as follows:
$s = - \frac{1}{P} \sum_{p=1}^{P} H(g_p),$
where $H$ represents entropy, and $g_{p}$ is the GLCM~\cite{haralick1973textural} derived from local patch $p$ in the original image with 64 gray levels, defined by radius $\delta=[1,2,3,4]$ and orientation $\theta=[0^{\circ}, 45^{\circ}, 90^{\circ}, 135^{\circ}]$.
In practice, we divide the gray image into local patches of size 64, and compute the average GLCM Score based on the partitioned local patches. 
Regrading the Compression Ratio, it is calculated as the ratio of the original image size in memory to the compressed image size using the JPEG algorithm at a quality of 95.
Qualitative and quantitative assessments constitute the comprehensive benchmark, Aesthetic-4K.

\subsection{Wavelet-based Fine-tuning}
\label{sec:wavelet}

In this section, we propose a \textbf{W}ave\textbf{L}et-based \textbf{F}ine-tuning (WLF) method suitable for various latent diffusion models, enabling  direct training with photorealistic images at $4096 \times 4096$ resolution. 
The core improvements consist of two components:

\noindent\textbf{Partitioned VAE.}  
The most commonly used VAEs~\cite{esser2024scaling, Flux:2024:Online} with $F=8$ encounter out-of-memory (OOM) issues during direct training and inference at $4096 \times 4096$.
To address this issue, we propose an efficient partitioned VAE, a simple yet effective method that compresses images with $F=16$, significantly reducing memory consumption.
Specifically, we use a dilation rate of 2 in the first convolutional layer of the VAE's encoder. 
In the last convolutional layer of the VAE's decoder, we partition the input feature map, up-sample by a factor of 2, apply the same convolution operator to each partitioned feature map, and then reorganize the final output.
Notably, our method maintains consistency in the latent space of pre-trained latent diffusion models, eliminating the need for retraining or fine-tuning VAEs, which prevents distribution shifts in the latent space and ensures compatibility with various diffusion models.

\noindent\textbf{Wavelet-based Latent Enhancement.}
Wavelet transform has achieved significant success in image processing and is widely used to decompose low-frequency approximations and high-frequency details from images or features~\cite{guth2022wavelet, phung2023wavelet}.
In this section, we propose wavelet-based latent enhancement, which emphasizes high-frequency components while preserving low-frequency information, thereby significantly enhancing fine details and rich textures in 4K image generation.
Consider the diffusion process, \ie
\begin{equation}
  \bm{z}_t = \alpha_t \cdot \bm{x}_0 + \sigma_t \cdot \epsilon,
  \label{eq:diffusion}
\end{equation}
where $\bm{x}_0$ and $\epsilon$ represent data distribution and standard normal distribution, respectively, and $\alpha_t$ and $\sigma_t$ are hyper-parameters in the diffusion formulation.
The original training objective in latent diffusion models is noise prediction~\cite{ho2020denoising, rombach2022high}, defined as:
\begin{equation}
  \epsilon_\Theta(\bm{z}_t, t) = \epsilon_t,
  \label{eq:noise_prediction}
\end{equation}
where $\Theta$ denotes the neural network, \eg, convolutional UNet or DiT.
Recent approaches, such as SD3~\cite{esser2024scaling} and Flux~\cite{Flux:2024:Online}, adopt rectified flows to predict velocity $\bm{v}$ parameterized by $\Theta$, with the objective as follows:
\begin{equation}
  \bm{v}_\Theta(\bm{z}_t, t) = \epsilon - \bm{x}_0.
  \label{eq:rectified_flow}
\end{equation}
Our method decomposes the low-frequency approximation and high-frequency details of latent features using wavelet transform, defining the training objective as follows:
\begin{equation}
  \mathcal{L}_{WLF} = \mathbb{E} \left[ w_t \Vert f(\bm{v}_\Theta(\bm{z}_t, t)) - f(\epsilon - \bm{x}_0) \Vert^2 \right],
  \label{eq:wavelet_rectified_flow}
\end{equation}
where $w_t$ is the loss weight, and $f(\cdot)$ denotes discrete wavelet transform (DWT).
Notably, we use the Haar wavelet, widely adopted in real-world applications due to its efficiency.
Specifically, $L= \frac{1}{\sqrt{2}}\left[1, 1\right]$ and $H= \frac{1}{\sqrt{2}}\left[-1, 1\right]$ denote the low-pass and high-pass filters, used to construct four kernels in DWT with stride $2$, namely $LL^T, LH^T, HL^T, HH^T$.
The DWT kernels are then used to decompose the input features into four sub-bands, the low-frequency approximation $\bm{x}_{ll}$ and high-frequency components $\bm{x}_{lh}, \bm{x}_{hl}, \bm{x}_{hh}$.
As a result, both low-frequency information and high-frequency details are incorporated in \cref{eq:wavelet_rectified_flow}, contributing to a comprehensive optimization in 4K image synthesis. 
In addition, our method supports various diffusion models by simply substituting the reconstruction objective, enabling seamless integration with conventional noise prediction approaches.

\section{Experiments}

To demonstrate the effectiveness of our method, we conduct experiments with the state-of-the-art latent diffusion models at various scales, including open-source SD3-2B~\cite{esser2024scaling} and Flux-12B~\cite{Flux:2024:Online}.
Specifically, mainstream evaluation metrics, such as FID~\cite{heusel2017gans}, Aesthetics~\cite{schuhmann2022laion} and CLIPScore~\cite{hessel2021clipscore}, as well as the proposed GLCM Score and Compression Ratio metrics, are reported for comprehensive assessments.
Additionally, we present quantitative and qualitative results that showcase the high-resolution image reconstruction and generation capabilities of partitioned VAE and WLF, respectively.

\subsection{Experimental Settings}

\noindent\textbf{Implementation Details.} 
Regrading the VAEs in both SD3~\cite{esser2024scaling} and Flux~\cite{Flux:2024:Online}, we encapsulate them using partitioned VAE with $F=16$, ensuring the direct training at $4096 \times 4096$ resolution without OOM issues. 
Notably, WLF is used for fine-tuning diffusion models, while the partitioned VAE and text encoder in pre-trained models are kept frozen, significantly improving the training efficiency.
In practice, we use the AdamW~\cite{loshchilov2017decoupled} optimizer with a constant learning rate of 1e-6 and weight decay of 1e-4. 
We employ mixed-precision training with a batch size of 32 and use ZeRO Stage 2 with CPU offload techniques~\cite{rajbhandari2020zero, ren2021zero}. 
The fine-tuning of SD3-2B and Flux-12B is conducted on 2 A800-80G GPUs and 8 A100-80G GPUs, respectively.

\subsection{Experiment Results}

\begin{figure}
  \centering
  \includegraphics[width=\linewidth]{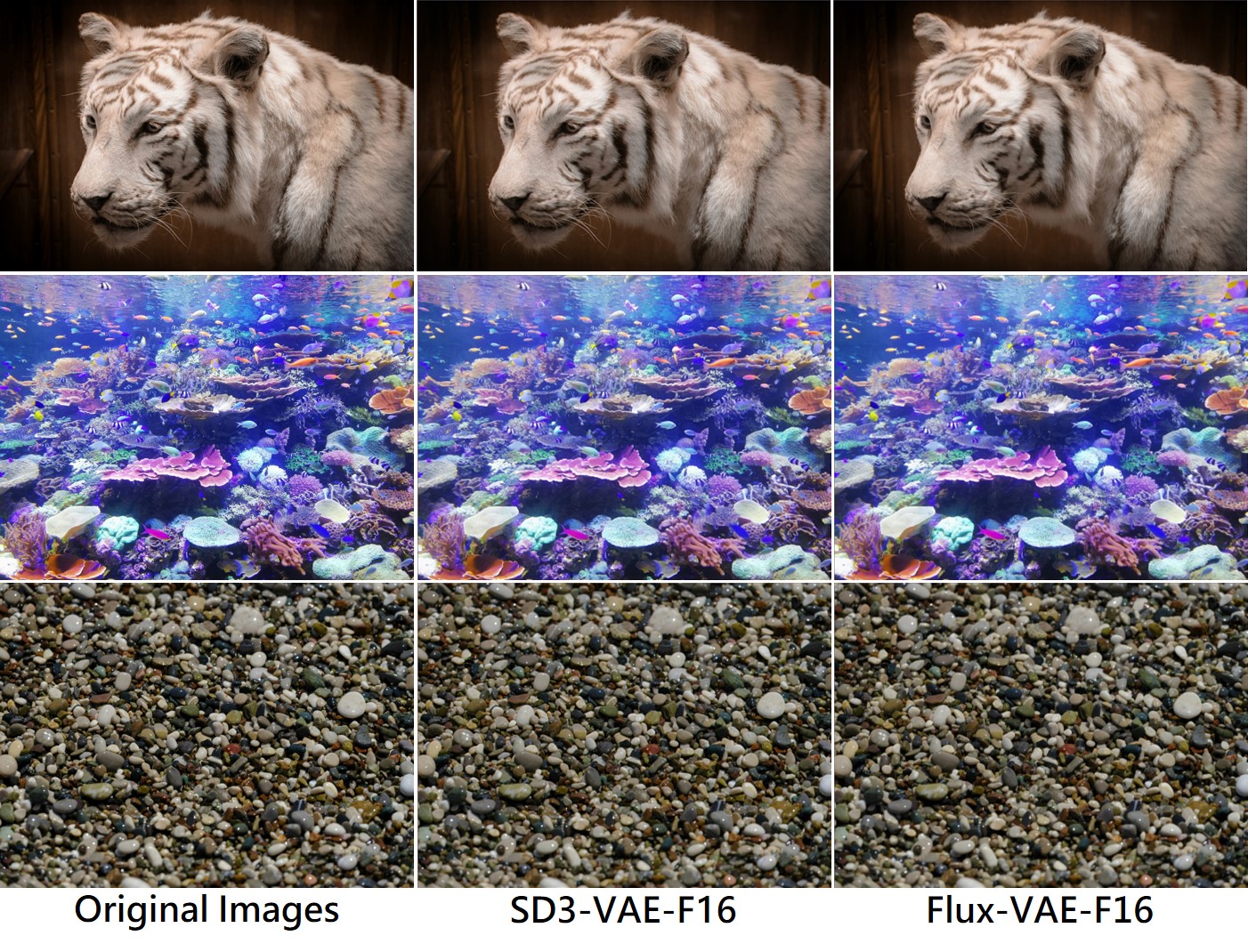}
  \caption{Reconstruction results of 4K images with partitioned VAEs of $F=16$.}
  \label{fig:vae_reconstruction}
\end{figure}

\noindent\textbf{Analysis of partitioned VAEs.}
In \cref{tab:vae_reconstruction}, we report the evaluation results, including rFID, Normalized Mean Square Error (NMSE), Peak Signal-to-Noise Ratio (PSNR), Structural Similarity Index Measure (SSIM), and LPIPS~\cite{zhang2018unreasonable}, to assess the reconstruction ability of partitioned VAEs on Aesthetic-4K at $4096 \times 4096$ resolution. 
We provide detailed results for VAEs in SD3 and Flux, enhanced by our partition techniques with $F=16$.
Notably, our partitioned VAE resolves the OOM issue encountered by the original VAE without requiring retraining or fine-tuning, thereby avoiding potential distribution shifts in the latent space.
In addition, we present visualizations of original and reconstruction images in \cref{fig:vae_reconstruction}.
The qualitative results demonstrate the effectiveness of the partitioned VAEs in 4K image reconstruction, ensuring consistency in the latent space for subsequent fine-tuning with 4K images.

\begin{table}
  \centering
  \resizebox{.45\textwidth}{!}{
  \begin{tabular}{l|ccccc}
    \toprule
    Model & rFID & NMSE & PSNR & SSIM & LPIPS \\
    \midrule
    SD3-VAE-F16 & 1.40 & 0.09 & 28.82 & 0.76 & 0.15 \\ 
    Flux-VAE-F16 & 1.69 & 0.08 & 29.22 & 0.79 & 0.16 \\ 
    \bottomrule
  \end{tabular}
  }
  \caption{Quantitative reconstruction results of VAEs with down-sampling factor of $F=16$ on our Aesthetic-4K training set at $4096 \times 4096$.}
  \label{tab:vae_reconstruction}
\end{table}

\noindent\textbf{Image Quality Assessment.}
Regarding image quality assessment, we conduct comprehensive comparisons using mainstream evaluation metrics, such as FID~\cite{heusel2017gans}, Aesthetics~\cite{schuhmann2022laion} and CLIPScore~\cite{hessel2021clipscore}, to provide an intuitive understanding of image quality and text prompt adherence.
In \cref{tab:eval_2048}, we report experimental results with SD3-2B and Flux-12B on Aesthetic-Eval@2048 in \cref{tab:eval_2048}, demonstrating the effectiveness of our method.

As aforementioned, these conventional evaluation metrics are insufficient for comprehensive assessment in ultra-high-resolution image synthesis, particularly in evaluating the fine details of 4K images. 
In addition to commonly used benchmarks, we also compare the GLCM Score, which aims to assess the texture richness in ultra-high-resolution images. 
Simultaneously, we also report the Compression Ratio using the JPEG algorithm at a quality setting of 95, which can serve as an important reference for assessing image quality with fine details.
As depicted in \cref{tab:eval_2048_glcm}, experimental results highlight the advantages of WLF in these metrics, distinctly demonstrating our method's ability to generate 4K images with fine details.

\begin{table}
  \centering
  \resizebox{.45\textwidth}{!}{
  \begin{tabular}{l|ccc}
    \toprule
    Model & FID $\downarrow$ & CLIPScore $\uparrow$ & Aesthetics $\uparrow$ \\ 
    \midrule
    SD3-F16@2048 & 43.82 & 31.50 & 5.91 \\ 
    SD3-F16-WLF@2048 & \textbf{40.18} & \textbf{34.04} & \textbf{5.96} \\ 
    \midrule
    Flux-F16@2048 & 50.57 & 30.41 & 6.36 \\ 
    Flux-F16-WLF@2048 & \textbf{39.49} & \textbf{34.41} & \textbf{6.37} \\ 
    \bottomrule
  \end{tabular}
  }
  \caption{Quantitative benchmarks of latent diffusion models on Aesthetic-Eval@2048 at $2048 \times 2048$ resolution.}
  \label{tab:eval_2048}
\end{table}

\begin{table}
  \centering
  \resizebox{.48\textwidth}{!}{
  \begin{tabular}{l|ccc}
    \toprule
    Model & GLCM Score $\uparrow$ & Compression Ratio $\downarrow$ \\
    \midrule
    SD3-F16@2048 & 0.75 & 11.23 \\ 
    SD3-F16-WLF@2048 & \textbf{0.79} & \textbf{10.51} \\ 
    \midrule
    Flux-F16@2048 & 0.58 & 14.80 \\  
    Flux-F16-WLF@2048 & \textbf{0.61} & \textbf{13.60} \\ 
    \bottomrule
  \end{tabular}
  }
  \caption{GLCM Score and Compression Ratio of latent diffusion models on Aesthetic-Eval@2048 at $2048 \times 2048$ resolution. } 
  \label{tab:eval_2048_glcm}
\end{table}

\begin{figure*}
  \centering
  \includegraphics[width=0.935\linewidth]{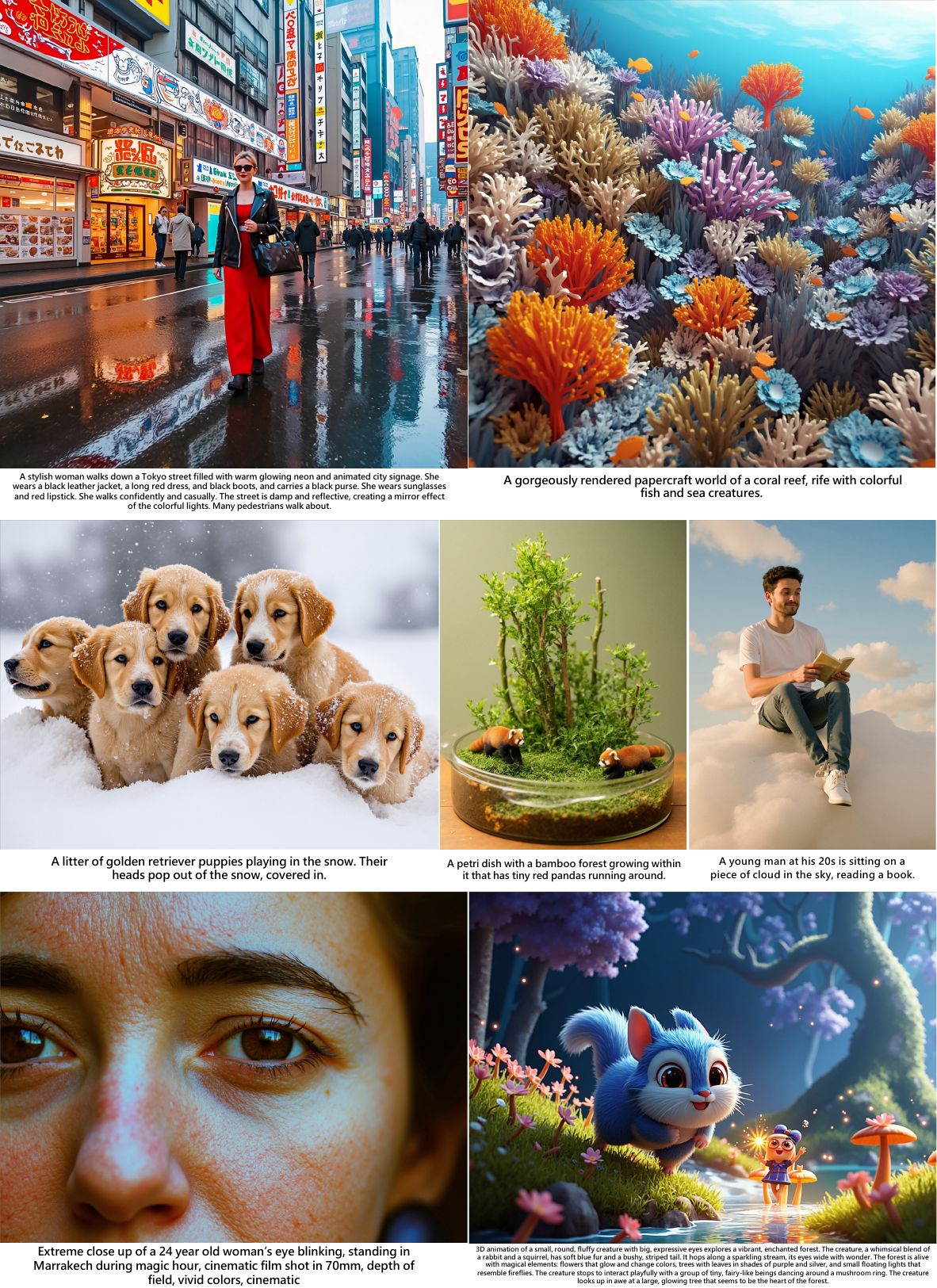}
  \caption{Qualitative 4K images synthesis of Diffusion-4K. Prompts are from Sora~\cite{Sora:2024:Online}.}
  \label{fig:qualitative_images}
\end{figure*}

\noindent\textbf{Qualitative 4K Image Synthesis.}
As illustrated in ~\cref{fig:qualitative_images}, we present qualitative high-resolution images synthesized with Diffusion-4K, powered by the state-of-the-art latent diffusion model, Flux-12B. 
Although WLF fine-tunes at $4096 \times 4096$ resolution, our method can synthesize ultra-high-resolution images at various aspect ratios, and even supports direct photorealistic image generation at higher resolutions.  
In addition, we report the inference details in \cref{tab:memory_and_speed}, highlighting the time and memory consumption of our method for directly generating 4K images.
For sampling, images are generated by discretizing the ordinary differential equation (ODE) process using an Euler solver. 
Please refer to the supplementary materials for more comparisons and details.

\begin{table}
  \centering
  \resizebox{.48\textwidth}{!}{
  \begin{tabular}{l|cccc}
    \toprule
    Model & Memory & Time (s/step) \\
    \midrule
    SD3-F8@4096 & OOM  & -  \\ 
    SD3-F16-WLF@4096  & 31.3GB  & 1.16  \\
    SD3-F16-WLF@4096 (CPU offload) & 16.1GB  & 1.22  \\
    \midrule
    Flux-F8@4096 & OOM & -  \\ 
    Flux-F16-WLF@4096  & 50.4 GB & 2.42  \\
    Flux-F16-WLF@4096 (CPU offload)  & 26.9 GB & 3.16  \\ 
    \bottomrule
  \end{tabular}
  }
  \caption{Memory consumption and inference speed of direct image synthesis at $4096 \times 4096$. The result is tested on one A100 GPU with BF16 Precision. }
  \label{tab:memory_and_speed}
\end{table}

\noindent\textbf{Preference Study.}
Human preference is assessed by rating the pairwise outputs from two models, \ie Flux with and without WLF.
Additionally, we employ the advanced multi-modal model, GPT-4o~\cite{hurst2024gpt}, as the evaluator for the preference study.
The following questions are asked:

\noindent\textbf{\emph{Visual Aesthetics:}} \emph{Given the prompt, which image is of higher-quality and aesthetically more pleasing?}

\noindent\textbf{\emph{Prompt Adherence:}} \emph{Which image looks more representative to the text shown above and faithfully follows it?}

\noindent\textbf{\emph{Fine Details:}} \emph{Which image more accurately represents the fine visual details? Focus on clarity, sharpness, and texture. Assess the fidelity of fine elements such as edges, patterns, and nuances in color. The more precise representation of these details is preferred! Ignore other aspects.}

\begin{figure}
  \centering
  \includegraphics[width=0.92\linewidth]{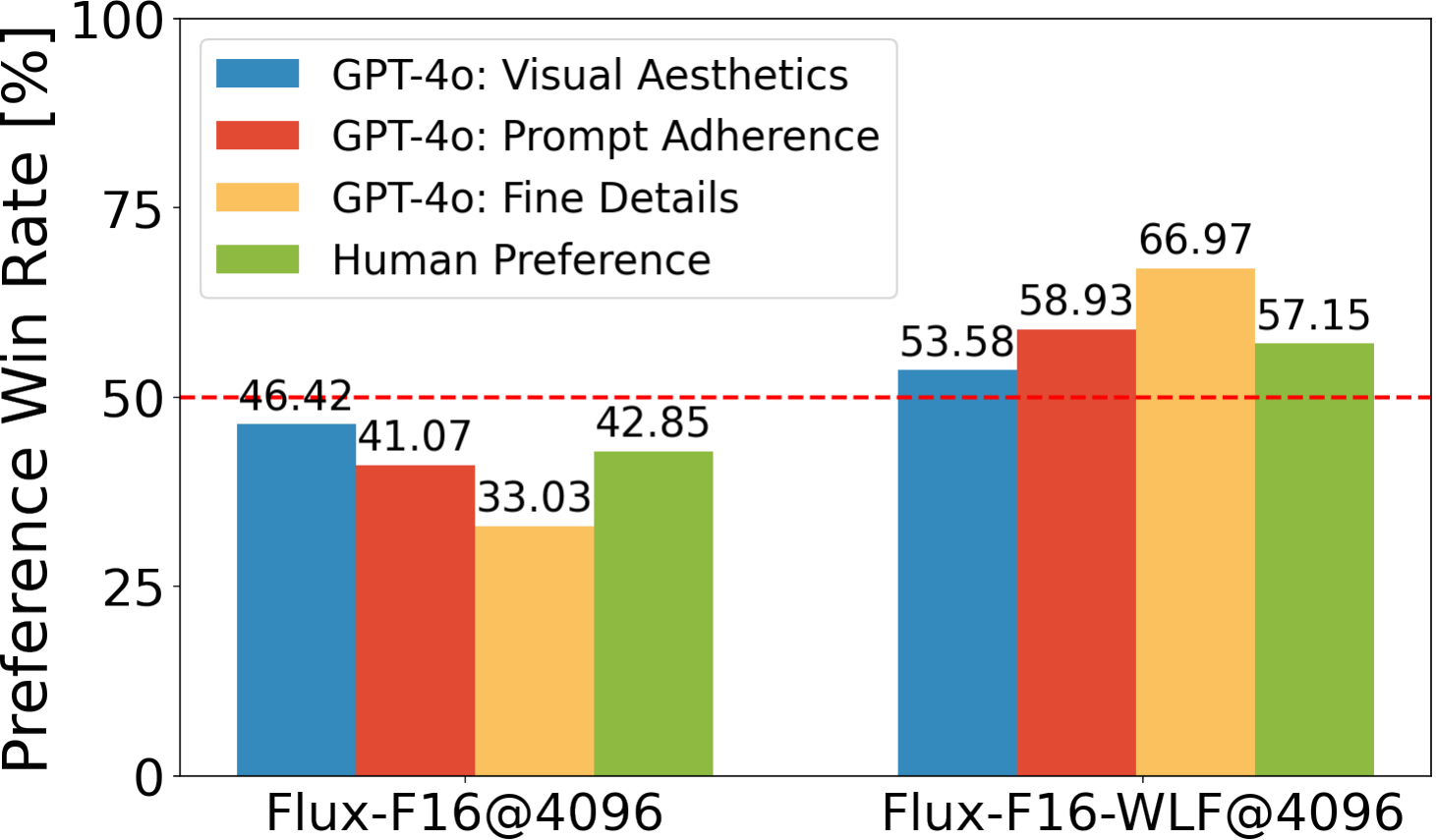}
  \caption{Human and GPT-4o preference evaluation. }
  \label{fig:win_rate}
\end{figure}

As shown in \cref{fig:win_rate}, our method demonstrates a superior win rate in both human and AI preference, highlighting the effectiveness of our WLF method in various aspects, including visual aesthetics, prompt adherence and fine details performance in 4K images.

\subsection{Ablation Studies}
\noindent\textbf{Ablation on Quality of Image Captions.}
We compare 4K image synthesis using original captions from LAION-5B~\cite{schuhmann2022laion} and captions generated by GPT-4o.
As shown in \cref{tab:ablation_captions}, both SD3 and Flux exhibit enhanced results in Aesthetic-Eval@4096 with captions from GPT-4o.
Quantitative results indicate that prompts generated by GPT-4o improve image synthesis and prompt coherence, demonstrating the significance of high-quality prompts in 4K image generation.

\begin{table}
  \centering
  \resizebox{.45\textwidth}{!}{
  \begin{tabular}{lc|cc}
    \toprule
    Captions & Model & CLIPScore $\uparrow$ & Aesthetics $\uparrow$  \\ 
    \midrule
    LAION-5B & SD3-F16@4096 & 29.37 & 5.90 \\ 
    GPT-4o & SD3-F16@4096 & \textbf{33.12} & \textbf{5.97} \\ 
    \midrule
    LAION-5B & Flux-F16@4096 & 29.12 & 6.02 \\ 
    GPT-4o & Flux-F16@4096 & \textbf{33.67} & \textbf{6.11} \\ 
    \bottomrule
  \end{tabular}
  }
  \caption{Ablation study on quality of image captions on Aesthetic-Eval@4096 at $4096 \times 4096$. }
  \label{tab:ablation_captions}
\end{table}

\noindent\textbf{Ablation on WLF.}
To demonstrate the effectiveness of WLF, we conduct ablation studies with SD3, comparing fine-tuning diffusion models with and without WLF, and report the experiment results of Aesthetic-Eval@4096 in \cref{tab:ablation_wlf} and \cref{tab:ablation_wlf_glcm}.
Compared to fine-tuning without WLF, our WLF method demonstrates superior performance in CLIPScore, Aesthetics, GLCM Score and Compression Ratio, significantly showcasing its effectiveness in visual aesthetics, prompt adherence, and high-frequency details.

\begin{table}
  \centering
  \resizebox{.45\textwidth}{!}{
  \begin{tabular}{l|cc}
    \toprule
    Model & CLIPScore $\uparrow$ & Aesthetics $\uparrow$  \\ 
    \midrule
    SD3-F16@4096 & 33.12 & 5.97 \\
    SD3-F16-finetune@4096 & 34.14 & 5.99  \\ 
    SD3-F16-WLF@4096 & \textbf{34.40} & \textbf{6.07}  \\ 
    \bottomrule
  \end{tabular}
  }
  \caption{Ablation on CLIPScore and Aesthetics of SD3 on Aesthetic-Eval@4096 at $4096 \times 4096$. SD3-F16-finetune@4096 represents fine-tuning without WLF.}
  \label{tab:ablation_wlf}
\end{table}

\begin{table}
  \centering
  \resizebox{.48\textwidth}{!}{
  \begin{tabular}{l|cc}
    \toprule
    Model & GLCM Score $\uparrow$ & Compression Ratio $\downarrow$  \\ 
    \midrule
    SD3-F16@4096 & 0.73 & 11.97 \\
    SD3-F16-finetune@4096 & 0.74 & 11.41 \\ 
    SD3-F16-WLF@4096 & \textbf{0.77} & \textbf{10.50} \\  
    \bottomrule
  \end{tabular}
  }
  \caption{Ablation on GLCM Score and Compression Ratio of SD3 on Aesthetic-Eval@4096 at $4096 \times 4096$. SD3-F16-finetune@4096 denotes fine-tuning without WLF. } 
  \label{tab:ablation_wlf_glcm}
\end{table}

\section{Conclusion}

In this paper, we present Diffusion-4K, a novel framework for ultra-high-resolution text-to-image generation. 
We introduce Aesthetic-4K benchmark and wavelet-based fine-tuning, capable of training with the state-of-the-art latent diffusion models, such as SD3 and Flux.
Qualitative and quantitative results demonstrate the effectiveness of our approach in training and generating photorealistic 4K images.

\section{Acknowledgement}

This work is partly supported by the National Natural Science Foundation of China (82441024), the Research Program of State Key Laboratory of Critical Software Environment, and the Fundamental Research Funds for the Central Universities.

{
    \small
    \bibliographystyle{ieeenat_fullname}
    \bibliography{main}
}


\clearpage
\setcounter{page}{1}

\twocolumn[{
\renewcommand\twocolumn[1][]{#1}
\maketitlesupplementary
\begin{center}
\centering
\captionsetup{type=figure}
\includegraphics[width=\linewidth]{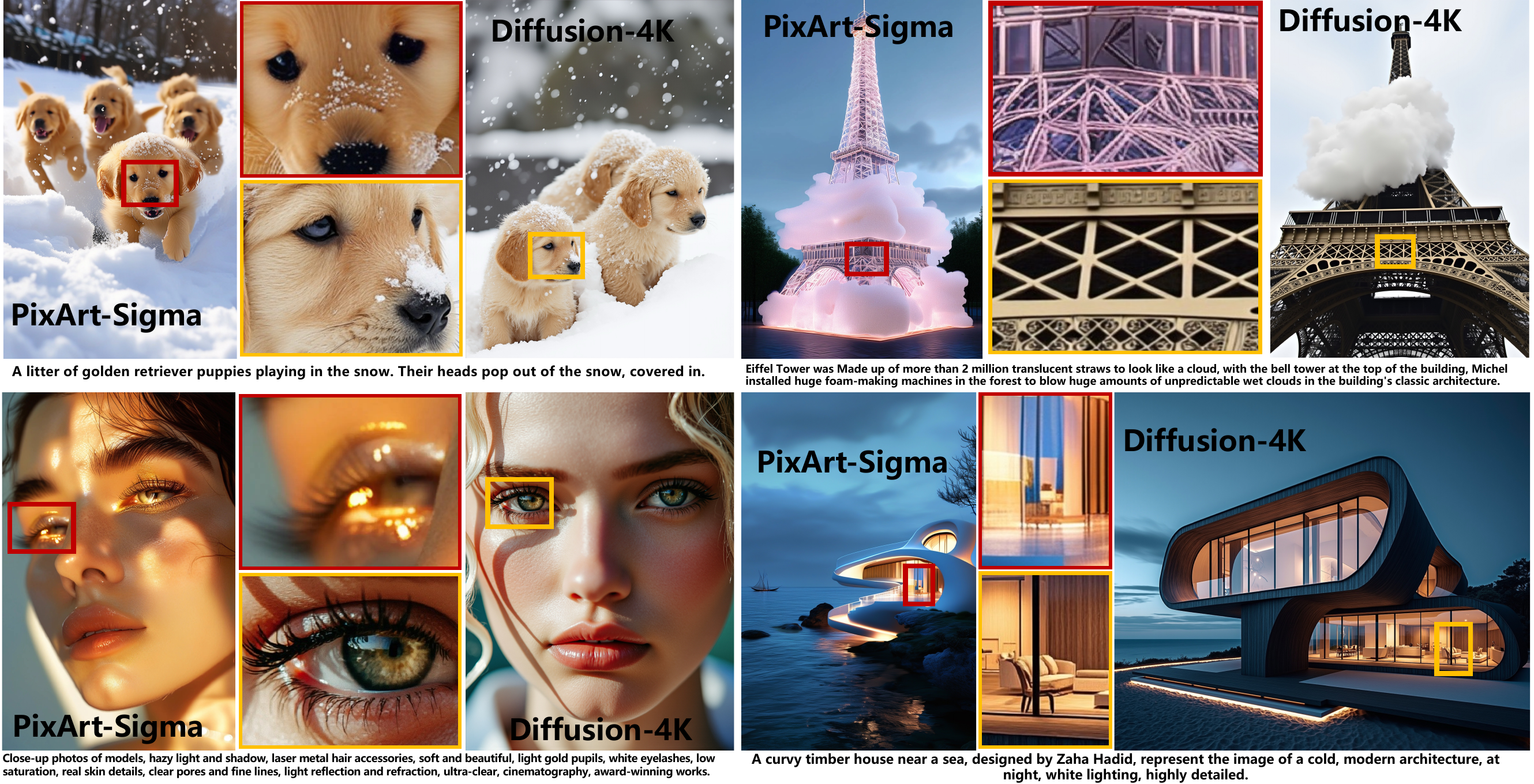}
\captionof{figure}{
We present comparisons with PixArt-$\Sigma$~\cite{chen2024pixart} using identical prompts, with images from PixArt-$\Sigma$ displayed on the left and those synthesized by our Diffusion-4K  shown on the right.
Our approach demonstrates significant superiority over PixArt-$\Sigma$ in fine details, as evidenced by the yellow patches \vs the red patches.
}
\label{fig:comparisons_supplementary}
\end{center}
}]

In the supplementary material, we provide additional details omitted from the main paper due to space constraints. 

\section{Comparisons}
\label{sec:comparisons}

\begin{figure*}[ht!]
  \centering
  \includegraphics[width=0.95\linewidth]{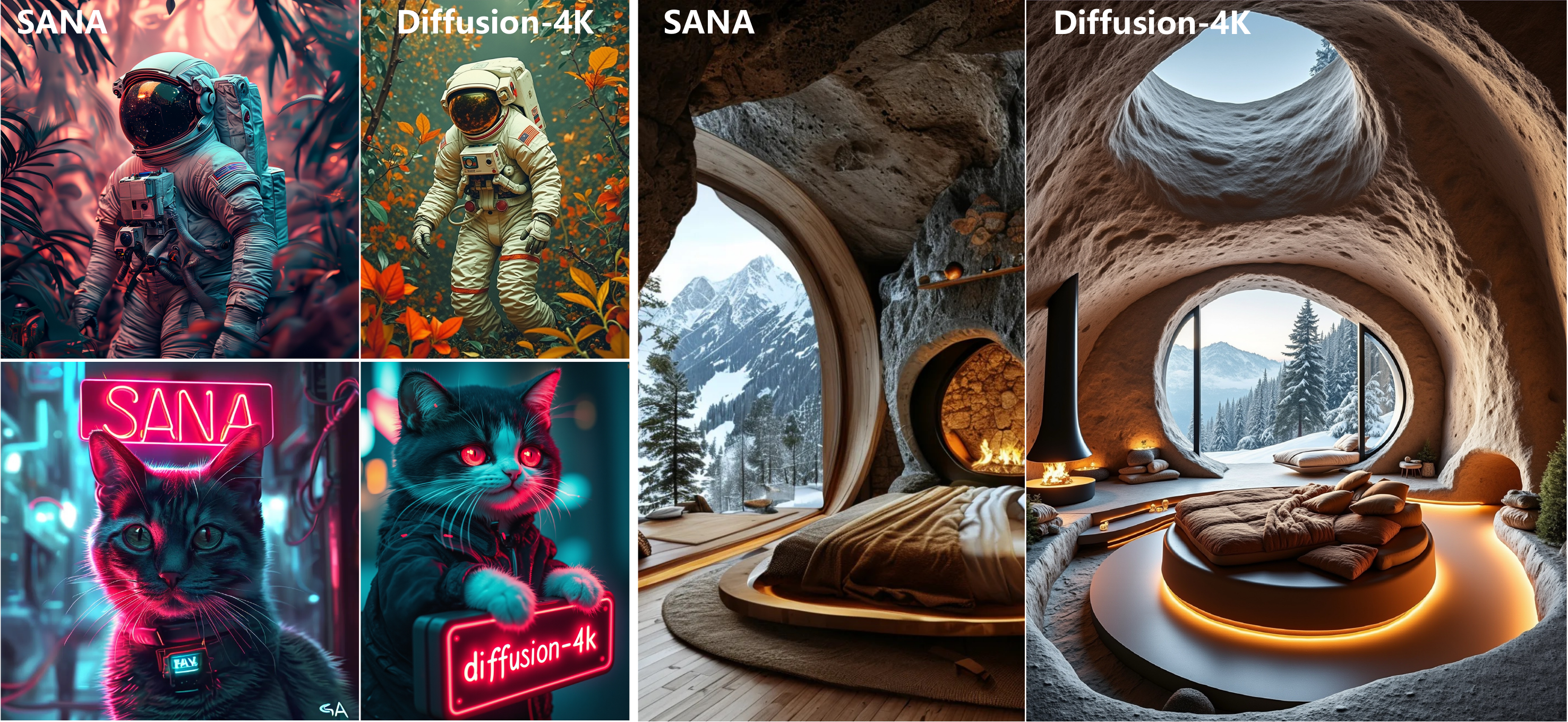}
  \caption{Comparisons with Sana~\cite{xie2024sana}.  }
  \label{fig:comparisons_sana_supplementary}
\end{figure*}

\noindent\textbf{Performance Comparisons.} 
We present detailed comparisons with other direct ultra-high-resolution image synthesis methods, including PixArt-$\Sigma$~\cite{chen2024pixart} and Sana~\cite{xie2024sana}.
As shown in \cref{fig:comparisons_supplementary}, our Diffusion-4K consistently produces impressive fine details in generated images under identical prompts, highlighting its superiority over PixArt-$\Sigma$~\cite{chen2024pixart} in 4K image generation. 
In addition, we compare the generated images of our Diffusion-4K with Sana~\cite{xie2024sana} in \cref{fig:comparisons_sana_supplementary}. 
Notably, PixArt-$\Sigma$ generates ultra-high-resolution images exclusively at a resolution of $3840 \times 2160$, and the images of PixArt-$\Sigma$ and Sana are sourced directly from their official websites.

\begin{figure*}[ht]
\centering
\includegraphics[width=0.95\linewidth]{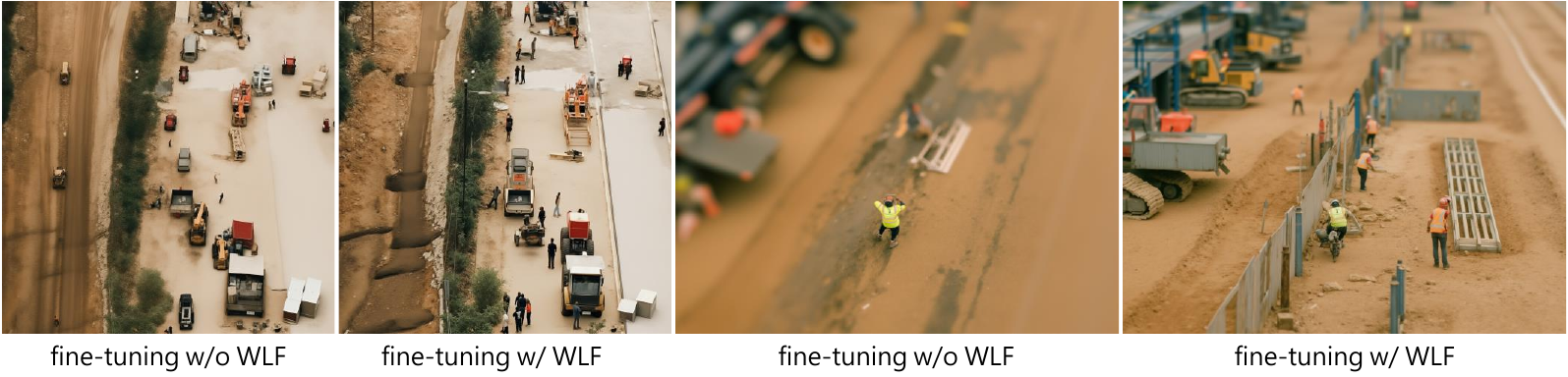} 
\caption{Ablation study on WLF. } 
\label{fig:ablation_wlf}
\end{figure*}

\noindent\textbf{Qualitative Evaluation of WLF.} 
As shown in \cref{eq:wavelet_rectified_flow}, WLF decomposes the latent into high- and low-frequency components, enabling the model to refine details (high-frequency) while preserving the overall structure (low-frequency).
This decomposition not only enhances the model's capability to generate fine details but also ensures that the changes don't disrupt the underlying patterns, making the fine-tuning process both efficient and precise. 
To comprehensively assess WLF, we additionally provide qualitative comparisons of latent fine-tuning with and without WLF to demonstrate its effectiveness. 
As illustrated in \cref{fig:ablation_wlf}, images generated with WLF exhibit richer details compared to those generated without WLF.

\begin{figure*}
  \centering
  \includegraphics[width=0.92\linewidth]{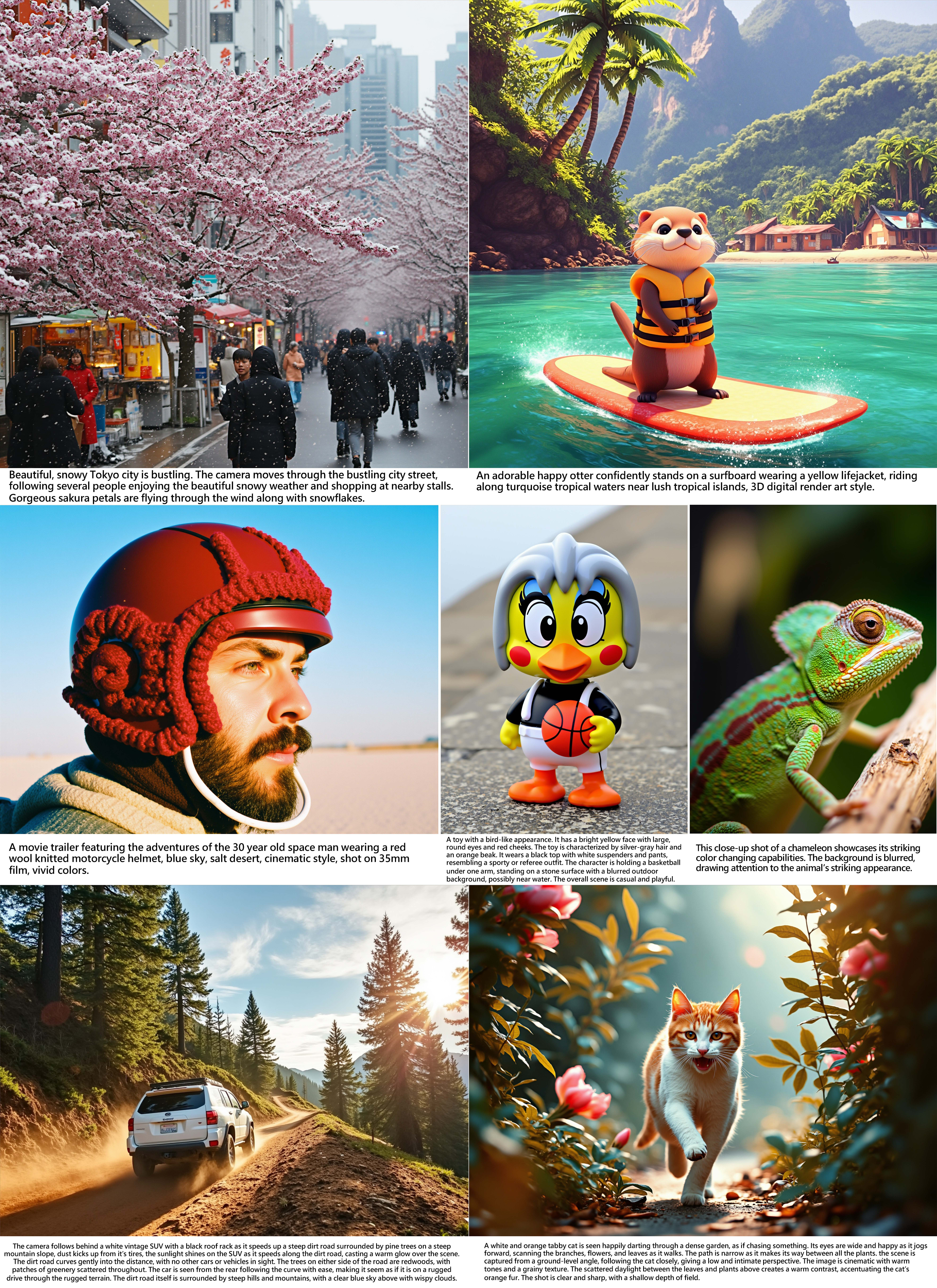}
  \caption{High-quality images synthesized by our Diffusion-4k. }
  \label{fig:generate_supplementary}
\end{figure*}

\begin{figure*}
  \centering
  \includegraphics[width=0.95\linewidth]{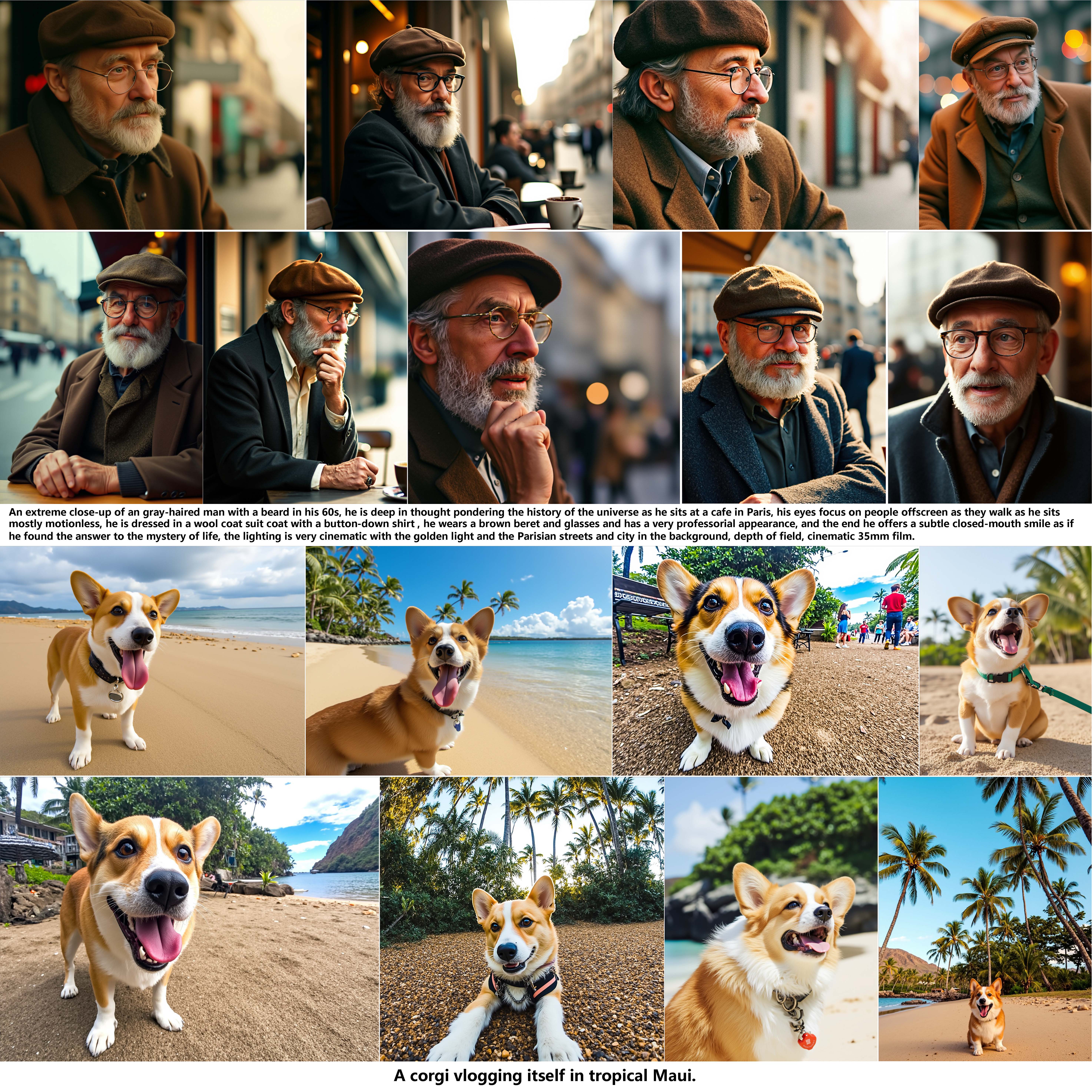}
  \caption{Synthesized images with different aspect ratios and random seeds. }
  \label{fig:generate_seeds_supplementary}
\end{figure*}

\begin{figure*}
  \centering
  \includegraphics[width=0.95\linewidth]{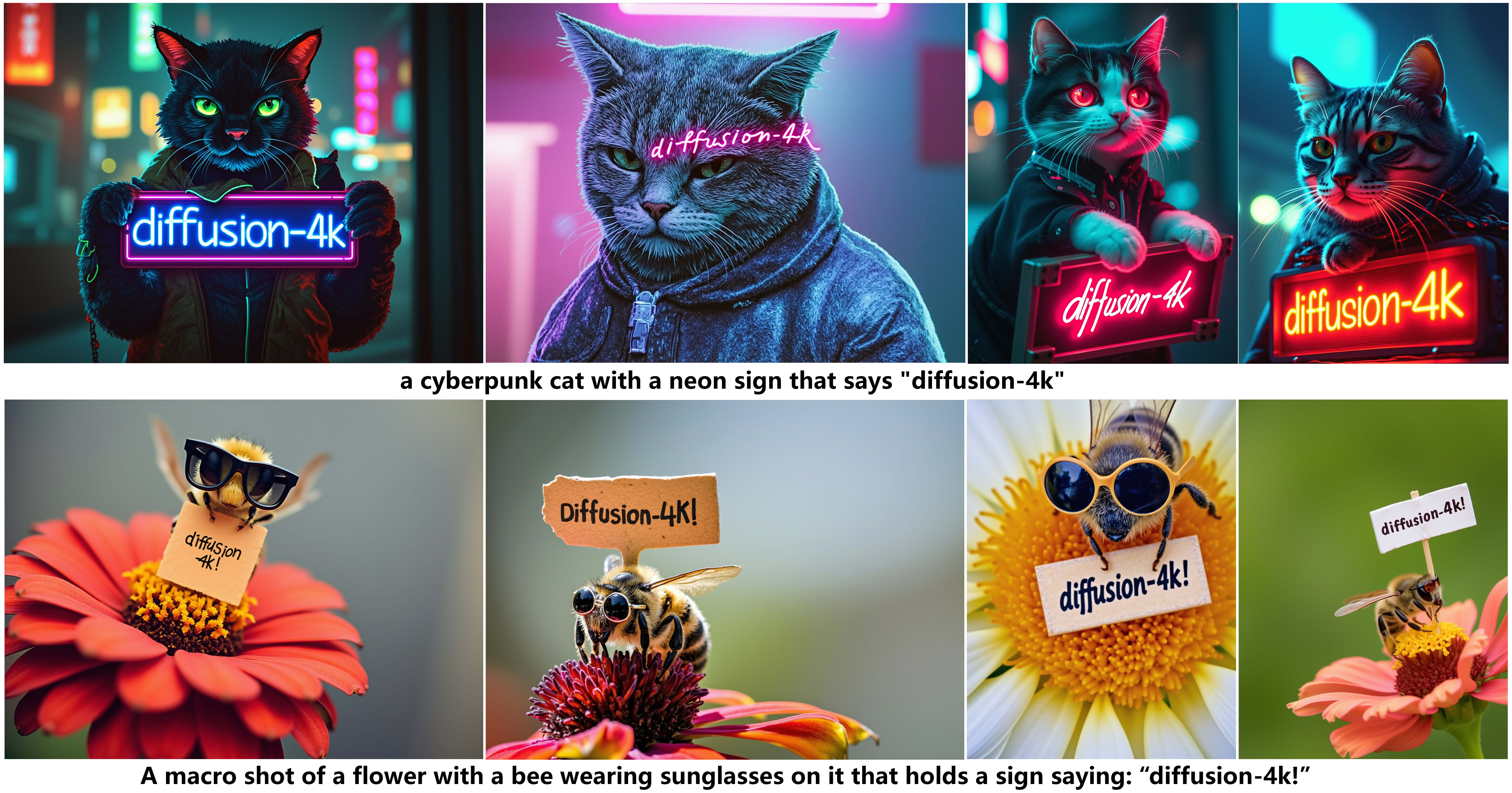}
  \caption{Synthesized images with spelled texts. }
  \label{fig:generate_text_supplementary}
\end{figure*}

\noindent\textbf{More Results of 4K Image Synthesis.} 
In \cref{fig:generate_supplementary}, we present the generated 4K images with different text prompts, demonstrating the effectiveness of our Diffusion-4K method in terms of visual aesthetics, adherence to text prompts and fine details. 
Additionally, as displayed in \cref{fig:generate_seeds_supplementary} and \cref{fig:generate_text_supplementary}, we provide the prompts and corresponding synthesized 4K images with varying aspect ratios, random seeds and spelled texts, illustrating diversity within the generated images. 
Qualitative results significantly demonstrate the impressive performance of our approach in 4K image generation, with particular attention to fine details.

\section{Aesthetic-4K Dataset}
\label{sec:aesthetic_4k_dataset}

\noindent\textbf{Details of Aesthetic-4K Dataset.} 
The LAION-Aesthetics dataset contains approximately $0.03\%$ 4K images, demonstrating the scarcity of 4K images in open-source datasets.
In the open-source PixArt-30k dataset~\cite{chen2024pixart}, the median height and width are 1615 and 1801 pixels, respectively.
In comparison, the Aesthetic-Train dataset has a median height and width of 4128 and 4640 pixels, respectively, marking a substantial improvement, as shown in \cref{tab:dataset}.
Additionally, we have meticulously filtered out low-quality images through manual inspection, excluding those with motion blur, focus issues, and mismatched text prompts, \etc.
We believe that constructing a high-quality training dataset at 4K resolution is one of the critical factors for 4K image generation.

\begin{table}[h]
\centering
\resizebox{\columnwidth}{!}{
\begin{tabular}{c|c|c|c|c}
\toprule
Dataset & Median height & Median width & Average height & Average width \\
\midrule
PixArt-30k  & 1615 & 1801 & 2531 & 2656 \\
Aesthetic-Train  & 4128 & 4640 & 4578 & 4838 \\
Aesthetic-Eval@2048 & 2983 & 3613 & 3143 & 3746 \\
Aesthetic-Eval@4096 & 4912 & 6449 & 5269 & 6420 \\
\bottomrule
\end{tabular}
}
\caption{Statistical comparisons of Aesthetic-4K and PixArt-30k.}
\label{tab:dataset}
\end{table}

\begin{figure}[ht!]
  \centering
  \includegraphics[width=\linewidth]{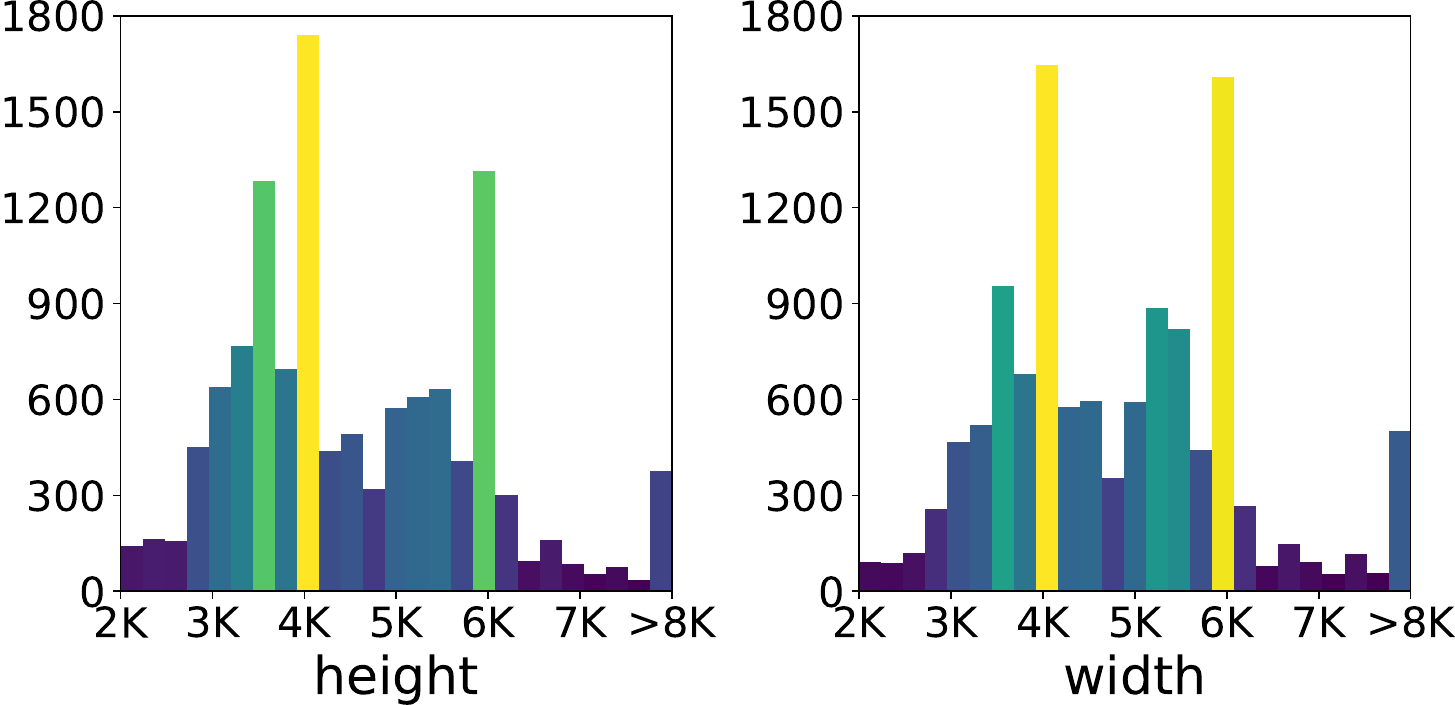}
  \caption{Histograms of image height and width in Aesthetic-4K. }
  \label{fig:aesthetic-4k-statistic}
\end{figure}

\begin{figure}[ht!]
  \centering
  \includegraphics[width=\linewidth]{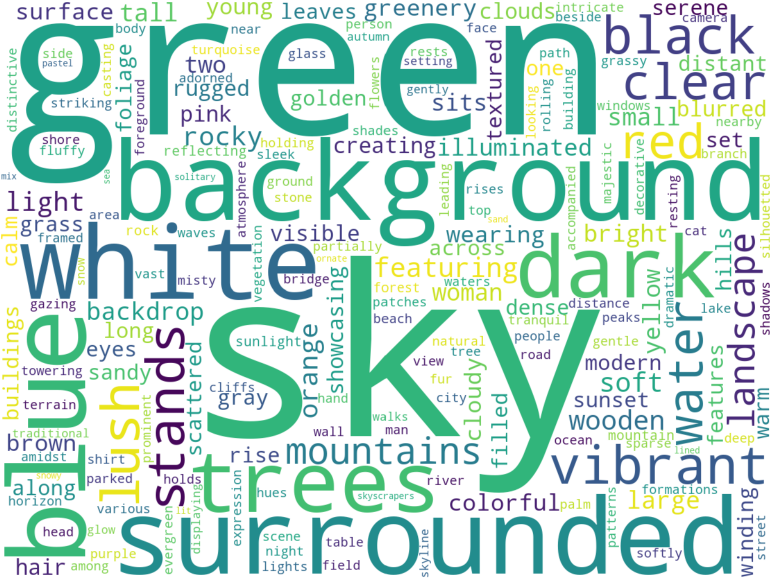}
  \caption{Word cloud of image captions from Aesthetic-4K. }
  \label{fig:aesthetic-4k-word-cloud}
\end{figure}

We provide the statistical histograms of image height and width for the Aesthetic-4K dataset in \cref{fig:aesthetic-4k-statistic}, highlighting the significant improvement in image resolution. 
In addition, we present a word cloud of image captions from the Aesthetic-4K dataset in \cref{fig:aesthetic-4k-word-cloud}, providing a visual representation of textual distribution.

\begin{figure*}
  \centering
  \includegraphics[width=0.92\linewidth]{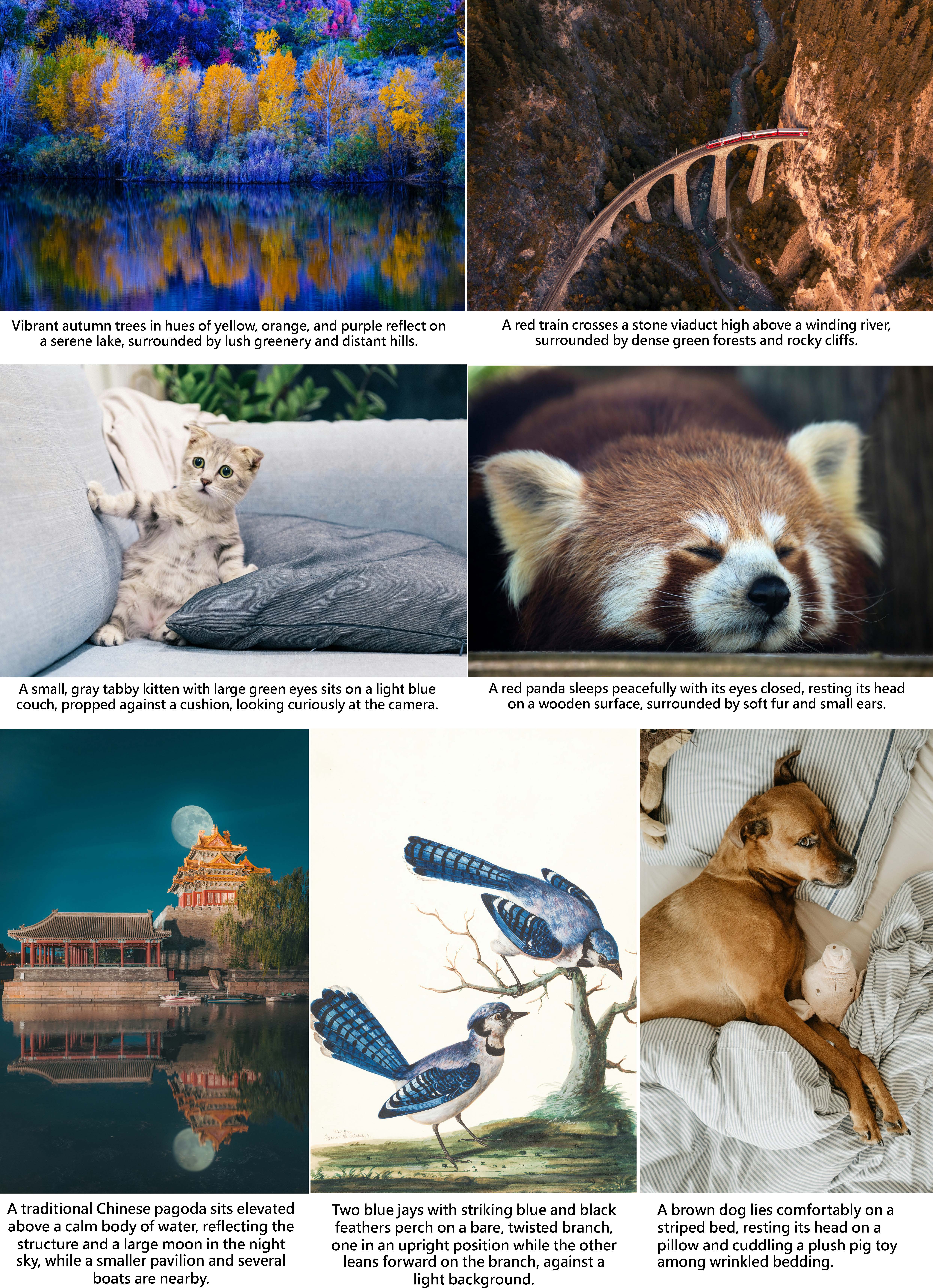}
  \caption{High-quality image-text samples in our Aesthetic-4K Dataset. }
  \label{fig:dataset_supplementary}
\end{figure*}

\noindent\textbf{Qualitative Samples in Aesthetic-4K Dataset.} 
As depicted in \cref{fig:dataset_supplementary}, we provide more image-text samples from the training set of our Aesthetic-4K dataset to illustrate its diversity and richness. 
As previously noted, the Aesthetic-4K dataset stands out due to its exceptional quality, presenting ultra-high-resolution images paired with precisely generated captions by GPT-4o~\cite{hurst2024gpt}.

\section{More Details}
\label{sec:more_details}
\noindent\textbf{Training Details of Diffusion-4K.} 
During image pre-processing, images are resized to a shorter dimension of $4096$, randomly cropped to a $4096 \times 4096$ resolution, and normalized with a mean and standard deviation of $0.5$. 
Our partitioned VAE compresses the pixel space $\mathbb{R}^{H \times W \times 3}$ into a latent space $\mathbb{R}^{\frac{H}{F} \times \frac{W}{F} \times C}$, where $F=16$. 
The encoded latents are normalized using the mean and standard deviation from the pretrained latent diffusion models, which are globally computed over a subset of the training data.
The latent diffusion models are then optimized using the wavelet-based latent fine-tuning objective in \cref{eq:wavelet_rectified_flow}. 
Regarding the text encoder, both CLIP~\cite{radford2021learning} and T5-XXL~\cite{raffel2020exploring} serve as the default models for text comprehension in SD3~\cite{esser2024scaling} and Flux~\cite{Flux:2024:Online}. 
To conserve memory, text embeddings for latent diffusion models are pre-computed, eliminating the need to load text encoders into the GPU during the training phase. 
We employ a default patch size of $P=2$ for DiTs, including SD3-2B and Flux-12B.
Latent diffusion models are optimized using the WLF objective with all parameters unfrozen, whereas text encoders and the partitioned VAE remain fixed during training. 
Additionally, as shown in \cref{tab:training_details}, we provide training details with SD3-2B and Flux-12B, including training steps and throughput, demonstrating the efficiency of our WLF method in handling scalable DiTs at ultra-high resolutions. 
Our approach requires approximately 2,000 A100 GPU hours to fine-tune a 12B diffusion model, demonstrating high computational efficiency while minimizing resource consumption.
Note that we use the open-source Flux.1-dev version, which is trained using guidance distillation, and adopt the default guidance scale of 3.5 for WLF.

\begin{table}
  \centering
  \resizebox{.45\textwidth}{!}{
  \begin{tabular}{l|c|c}
    \toprule
    Model & SD3-2B-WLF  & Flux-12B-WLF  \\ 
    \midrule
    Training steps & 20K & 20K \\ 
    Throughput (images/s) & 0.59 & 1.39 \\
    \bottomrule
  \end{tabular}
  }
  \caption{Training details of SD3-2B and Flux-12B with WLF at $4096 \times 4096$.}
  \label{tab:training_details}
\end{table}

\begin{table*}
  \centering
  \resizebox{.95\textwidth}{!}{
  \begin{tabular}{l|c}
    \toprule
    Tasks & Prompts \\ 
    \midrule
    \makecell[c]{Image \\ Caption} & \makecell[c]{ \{``\textbf{text}'': ``Directly describe with brevity and as brief as possible the scene or characters without any introductory phrase \\ like  `This image shows',  `In the scene',  `This image depicts' or similar phrases.  Just start describing the scene please.'' \} } \\
    \midrule
    \makecell[c]{Preference \\ Study} & \makecell[c]{ \{``\textbf{system}'': ``As an AI visual assistant, you are  analyzing two specific images.  When presented  with \\ a specific caption, it is required to  evaluate visual aesthetics,  prompt coherence  and fine details.'', \\ 
    ``\textbf{text}'': ``The caption for the two images is: $\langle$prompt$\rangle$.  Please answer the following questions: \\
    1. Visual Aesthetics: Given the prompt, which image is of  higher-quality and aesthetically more pleasing? \\
    2. Prompt Adherence: Which image looks more representative  to the text shown above and faithfully follows it? \\
    3. Fine Details: Which image more accurately represents the fine visual details? Focus on clarity, \\  sharpness, and texture.  Assess the fidelity of fine elements such as edges, patterns,  and nuances in color.  \\ The more precise representation of these details is preferred! Ignore other aspects. \\
    Please respond me strictly in the following format: \\
    1. Visual Aesthetics: $\langle$the first image is better$\rangle$ or  $\langle$the second image is better$\rangle$. The reason is $\langle$give your reason here$\rangle$. \\
    2. Prompt Adherence: $\langle$the first image is better$\rangle$ or $\langle$the second image is better$\rangle$. The reason is $\langle$give your reason here$\rangle$. \\
    3. Fine Details: $\langle$the first image is better$\rangle$ or  $\langle$the second image is better$\rangle$. The reason is $\langle$give your reason here$\rangle$. "\} }  \\
    \bottomrule
  \end{tabular}
  }
  \caption{Designed prompts for image caption and preference study with GPT-4o.}
  \label{tab:gpt_caption}
\end{table*}

\noindent\textbf{Detailed Prompts for GPT-4o.} 
In \cref{tab:gpt_caption}, we provide detailed prompts for image caption using GPT-4o, to generate precise text prompts for the Aesthetic-4K dataset.
Additionally, we present detailed prompts used in the preference study with GPT-4o, to evaluate AI preferences for generated images, including visual aesthetics, prompt adherence and fine details.

\begin{figure*}[ht!]
\centering
\includegraphics[width=0.95\textwidth]{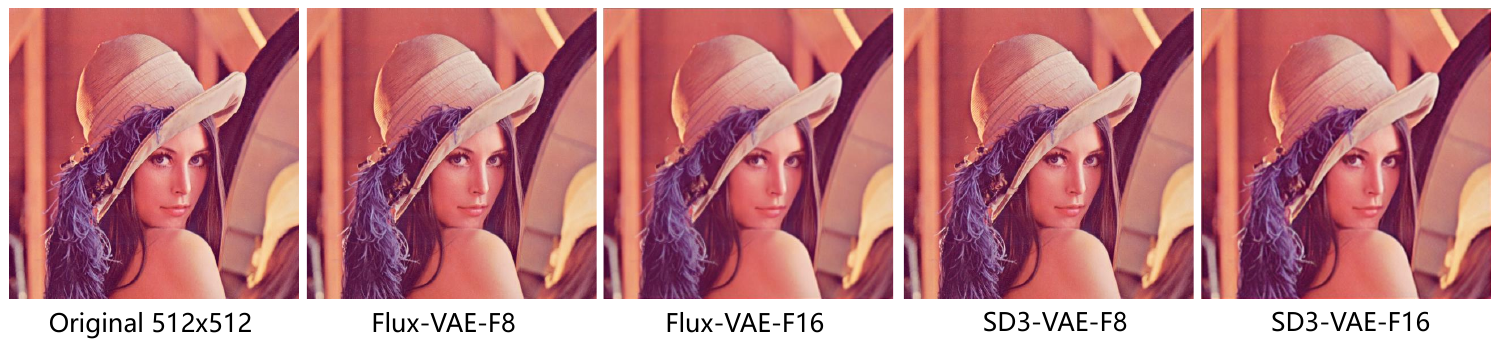} 
\caption{Reconstruction results of partitioned VAEs at $512 \times 512$. }
\label{fig:reconstrcution}
\end{figure*}

\noindent\textbf{Details for Evaluation.} 
During evaluation, images are generated using a guidance scale of 7.0, and flow-matching sampling introduced in SD3~\cite{esser2024scaling}, with 28 sampling steps for SD3-2B and 50 sampling steps for Flux-12B.
The FID~\cite{heusel2017gans}, Aesthetics~\cite{schuhmann2022laion} and CLIPScore~\cite{hessel2021clipscore} are computed using resized images at a fixed low resolution. 

To quantitatively analyze the alignment between our indicators and human ratings, five participants are asked to rate extracted patches on a scale from 1 to 10 based on visual details, with the average scores used to evaluate SRCC and PLCC as presented in \cref{tab:metrics}. 
Note that the SRCC and PLCC for the Compression Ratio are calculated using its reciprocal.

For human preference evaluation on 4K image synthesis in \cref{fig:win_rate}, we conduct experiment with 112 text prompts sampled from Sora~\cite{Sora:2024:Online}, PixArt~\cite{chen2024pixart}, SD3~\cite{esser2024scaling}, \etc. 
Ten participants are asked to rate the preference for the generated images in visual aesthetics, prompt adherence and fine details respectively.

\noindent\textbf{Reconstruction Results of Partitioned VAEs.} 
We provide the reconstruction results with images at a resolution of $512 \times 512$ in \cref{fig:reconstrcution}, showcasing the capability of partitioned VAEs in handling low-resolution images.

\end{document}